\documentclass{sig-alternate}

\usepackage[
pdftitle={Computing of Applied Digital Ecosystems},
pdfauthor={Gerard Briscoe},
pdfsubject={Digital Ecosystems},
pdfkeywords={Sustainability, Digital Ecosystems},
colorlinks=true, linkcolor=black, citecolor=black,  filecolor=black, urlcolor=black, hyperfootnotes=false]{hyperref} 

\usepackage{graphicx, setspace, acronym, ifthen, substr, color}

\usepackage[paper=a4paper, asymmetric, left=2cm,right=1.4cm,top=2cm,bottom=4.2cm]{geometry}

\newboolean{final}\setboolean{final}{true}

\newboolean{mactex}\setboolean{mactex}{true}

\newboolean{arxiv}\setboolean{arxiv}{false}

\ifthenelse{\boolean{final}}{}{\usepackage{showlabels}\usepackage{citeext}}

\ifthenelse{\boolean{arxiv}}
	{\newcommand{\arxivfootnote}{\footnote}}
	{\def\arxivfootnote#1{}}

\ifthenelse{\boolean{arxiv}}
	{\def\arxivOnly#1{#1}}
	{\def\arxivOnly#1{}}

\ifthenelse{\boolean{arxiv}}
	{\def\arxivNot#1{}}
	{\def\arxivNot#1{#1}}

\ifthenelse{\boolean{arxiv}}
	{}
	{}

\newcommand{\tfigure}[9]
        {
        \IfSubStringInString{!}{#7}{\begin{figure}[#7]}{\begin{figure}[!t]}
        \IfSubStringInString{mm}{#8}{\vspace{#8}}{}
        \centering
        
        \IfSubStringInString{pdf}{#3}
                {
                \ifthenelse{\boolean{mactex}}{}{\execute{cd images; ln -s #2.pdf .#2.gdf}}
	      \includegraphics[#1]{images/#2}
                }
                {
                \ifthenelse{\boolean{mactex}}{}{\execute{cd images; ./pdfcrop.sh #2}}
                \includegraphics[#1]{images/#2-crop.pdf}
                }

        \vspace{#6}
        \caption[#4]
                {
                \label{#2}
                #4: #5
                }
        \IfSubStringInString{mm}{#9}{\vspace{#9}}{}
        \end{figure}
        }

\newcommand{\Circlesub}[4]
	{
	\ifthenelse{\boolean{mactex}}{}{\immediate\write18{cd images; ./pdfcrop.sh circle#2}}
	\ifthenelse{\boolean{final}}
		{\hspace{#1}\raisebox{#4}{$\includegraphics[scale=0.5, clip=true, trim=0mm 0mm 0mm 0mm]{images/circle#2-crop.pdf}$}\hspace{#3}}
		{\href{file://localhost/Users/g/Desktop/PhDthesis/images/circle#2.graffle}{\hspace{#1}\raisebox{#4}{$\includegraphics[clip=true, trim=0mm 0mm 0mm 0mm]{images/circle#2-crop.pdf}$}\hspace{#3}}}
	}

\newcommand{\execute}[1]{\immediate\write18{#1}}

\definecolor{tred}{RGB}{255,0,0}

\newcommand{\setCap}[2]{#1\immediate\write18{./mkcaption.sh #2}}
\newcommand{\getCap}[1]{\acl*{#1}}

\acrodef{PCG}{Projected Conjugate Gradient} 
\acrodef{QP}{quadratic programming}
\acrodef{RBF}{Radial-Basis Function}
\acrodef{ABM}{Agent-Based Modelling}
\acrodef{AI}{Artificial Intelligence}
\acrodef{DAI}{Distributed Artificial Intelligence}
\acrodef{API}{Application Programming Interface}
\acrodef{ARF}{p14ARF human tumor-suppressor gene}
\acrodef{B2B}{business-to-business}
\acrodef{BDP}{Biological Design Pattern}
\acrodef{BGS}{Best Guess Solution}
\acrodef{BIC}{Biologically-Inspired Computing}
\acrodef{BML}{Business Modelling Language}
\acrodef{BPEL}{Business Process Execution Language}
\acrodef{BPMN}{Business Process Modelling Notation}
\acrodef{CAS}{Complex Adaptive Systems}
\acrodef{COBOL}{COmmon Business-Oriented Language}
\acrodef{DBE}{Digital Business Ecosystem}
\acrodef{DE}{Digital Ecosystem}
\acrodef{DEC}{distributed evolutionary computing}
\acrodef{DGA}{Distributed genetic algorithms}
\acrodef{DIS}{Distributed Intelligence System}
\acrodef{DNA}{Deoxyribose Nucleic Acid}
\acrodef{DOP}{DBE Open Protocol}
\acrodef{DSS}{Distributed Storage System}
\acrodef{EAP}{Evolving Agent Population}
\acrodef{ebXML}{e-business eXtensible Markup Language}
\acrodef{EC}{Evolutionary Computing}
\acrodef{ECJ}{Evolutionary Computing in Java}
\acrodef{EE}{Evolutionary Environment}
\acrodef{EFL}{Evolutionary Framework for Language}
\acrodef{FLE}{Framework for Language Ecosystems}
\acrodef{EOA}{Ecosystem-Oriented Architecture}
\acrodef{ESS}{evolutionary stable strategy}
\acrodef{EvE}{Evolutionary Environment}
\acrodef{ExE}{Execution Environment}
\acrodef{FCB}{Framework for Computational Biomimicry}
\acrodef{FFF}{Fitness Function Framework}
\acrodef{FL}{Fitness Landscape}
\acrodef{HWU}{Heriot-Watt University}
\acrodef{ICL}{Imperial College London}
\acrodef{ICT}{Information and Communications Technology}
\acrodef{INTEL}{Intel Ireland}
\acrodef{IPA}{International Phonetic Alphabet}
\acrodef{ISUFI}{Istituto Superiore Universitario di Formazione Interdisciplinare}
\acrodef{JDJ}{Java Developer's Journal}
\acrodef{KC}{Kolmogorov-Chaitin}
\acrodef{LAN}{local area network}
\acrodef{LSE}{London School of Economics and Political Science}
\acrodef{MAS}{Multi-Agent System}
\acrodef{MDL}{Minimum Description Length}
\acrodef{MDM2}{murine double minute 2}
\acrodef{MFT}{Mean Field Theory}
\acrodef{MoAS}{Mobile Agent System}
\acrodef{MOF}{Meta Object Facility}
\acrodef{MUH}{migration and usage history}
\acrodef{NIC}{Nature Inspired Computing}
\acrodef{NN}{Neural Network}
\acrodef{NoE}{Network of Excellence}
\acrodef{OMG}{Open Mac Grid}
\acrodef{OPAALS}{Open Philosophies for Associative Autopoietic Digital Ecosystems}
\acrodef{P2P}{peer-to-peer}
\acrodef{P53}{protein 53}
\acrodef{PDA}{Personal Digital Assistant}
\acrodef{QoS}{quality of service}
\acrodef{REST}{REpresentational State Transfer}
\acrodef{RNA}{Deoxyribose Nucleic Acid}
\acrodef{SAE}{Software Agent Ecosystem}
\acrodef{SBML}{Systems Biology Modelling Language}
\acrodef{SBVR}{Semantics of Business Vocabulary and Business Rules}
\acrodef{SDL}{Service Description Language}
\acrodef{SF}{Service Factory}
\acrodef{SIM}{Social Interaction Mechanism}
\acrodef{SM}{Service Manifest}
\acrodef{SME}{Small and Medium sized Enterprise}
\acrodef{SML}{Service Modelling Language}
\acrodef{SMO}{Sequential Minimal Optimisation}
\acrodef{SOA}{Service-Oriented Architecture}
\acrodef{SOAP}{Simple Object Access Protocol}
\acrodef{SOC}{Self-Organised Criticality}
\acrodef{SOLUTA}{SOLUTA.NET}
\acrodef{SOM}{Self-Organising Map}
\acrodef{SSL}{Semantic Service Language}
\acrodef{STU}{Salzburg Technical University}
\acrodef{SUN}{Sun Microsystems}
\acrodef{SVM}{Support Vector Machine}
\acrodef{TM}{Turing Machine}
\acrodef{UBHAM}{University of Birmingham}
\acrodef{UDDI}{Universal Description Discovery and Integration}
\acrodef{UML}{Unified Modelling Language}
\acrodef{URI}{Uniform Resource Identifier}
\acrodef{UTM}{Universal Turing Machine}
\acrodef{VLP}{variable length population}
\acrodef{VLS}{variable length sequences}
\acrodef{vls}{variable length sequence}
\acrodef{WP}{Work-Package}
\acrodef{WSDL}{Web Services Definition Language}
\acrodef{XMI}{XML Metadata Interchange}
\acrodef{XML}{eXtensible Markup Language}
\acrodef{MD5}{Message-Digest algorithm 5}
\acrodef{GA}{genetic algorithm}
\acrodef{GP}{genetic programming}
\acrodef{MASON}{Multi-Agent Simulator Of Neighbourhoods}
\acrodef{Repast}{Recursive Porous Agent Simulation Toolkit}
\acrodef{JCLEC}{Java Computing Library for Evolutionary Computing}
\acrodef{OWL-S}{Web Ontology Language - Service}
\acrodef{EGT}{Evolutionary Game Theory}
\acrodef{RBF}{Radial Basis Functions}
\acrodef{SWS}{Semantic Web Services}
\acrodef{HDD}{Hard Disk Drive}
\acrodef{SSD}{Solid-State Drive}

\acrodef{im1}{are different probabilities of going from island 1 to island 2, as there is of going from island 2 to island 1.}
\acrodef{im2}{mirrors the naturally inspired quality that although two populations have the same physical separation, it may be easier to migrate in one direction than the other, i.e. fish migration is easier downstream than upstream.}
\acrodef{digEco}{with the agents, the populations, the agent migration for \acl{DEC}, and the environmental selection pressures provided by the user base, then the union of the habitats creates the Digital Ecosystem}
\acrodef{archComTop}{many strongly connected clusters (communities), called {sub-networks} (quasi-complete graphs), with a few connections between these clusters (communities). Graphs with this topology have a very high clustering coefficient and small characteristic path lengths}
\acrodef{similarCap}{requests are evaluated on separate {islands} (populations), and so adaptation is accelerated by the sharing of solutions between evolving populations (islands), because they are working to solve similar requests (problems).}
\acrodef{picUser}{will formulate queries to the Digital Ecosystem by creating a request as a {semantic description}, like those being used and developed in \acp{SOA}}
\acrodef{picUserReq}{A population is then instantiated in the user's habitat in response to the user's request, seeded from the agents available at their habitat.}
\acrodef{urlCapUnifrom}{The {observed} frequencies of the application (agent aggregation) size mostly matched the {expected} frequencies}
\acrodef{urlCapGaussian}{The {observed} frequencies of the application (agent aggregation) size matched the {expected} frequencies with only minor variations}
\acrodef{urlpower}{The {observed} frequencies of the application (agent aggregation) size matched the {expected} frequencies with some variation}
\acrodef{urvunifromCap}{The {observed} frequencies for the number of agent attributes mostly matched the {expected} frequencies}
\acrodef{urvgaussianCap}{The {observed} frequencies for the number of agent attributes again followed the {expected} frequencies, but there was variation}
\acrodef{urvpowerCap}{The {observed} frequencies for the number of agent attributes also followed the {expected} frequencies, but there was variation}
\acrodef{im1}{are different probabilities of going from island 1 to island 2, as there is of going from island 2 to island 1.}
\acrodef{im2}{mirrors the naturally inspired quality that although two populations have the same physical separation, it may be easier to migrate in one direction than the other, i.e. fish migration is easier downstream than upstream.}
\acrodef{digEco}{with the agents, the populations, the agent migration for \acl{DEC}, and the environmental selection pressures provided by the user base, then the union of the habitats creates the Digital Ecosystem}
\acrodef{archComTop}{many strongly connected clusters (communities), called {sub-networks} (quasi-complete graphs), with a few connections between these clusters (communities). Graphs with this topology have a very high clustering coefficient and small characteristic path lengths}
\acrodef{similarCap}{requests are evaluated on separate {islands} (populations), and so adaptation is accelerated by the sharing of solutions between evolving populations (islands), because they are working to solve similar requests (problems).}
\acrodef{picUser}{will formulate queries to the Digital Ecosystem by creating a request as a {semantic description}, like those being used and developed in \acp{SOA}}
\acrodef{picUserReq}{A population is then instantiated in the user's habitat in response to the user's request, seeded from the agents available at their habitat.}
\acrodef{urlCapUnifrom}{The {observed} frequencies of the application (agent aggregation) size mostly matched the {expected} frequencies}
\acrodef{urlCapGaussian}{The {observed} frequencies of the application (agent aggregation) size matched the {expected} frequencies with only minor variations}
\acrodef{urlpower}{The {observed} frequencies of the application (agent aggregation) size matched the {expected} frequencies with some variation}
\acrodef{urvunifromCap}{The {observed} frequencies for the number of agent attributes mostly matched the {expected} frequencies}
\acrodef{urvgaussianCap}{The {observed} frequencies for the number of agent attributes again followed the {expected} frequencies, but there was variation}
\acrodef{urvpowerCap}{The {observed} frequencies for the number of agent attributes also followed the {expected} frequencies, but there was variation}
\acrodef{im1}{are different probabilities of going from island 1 to island 2, as there is of going from island 2 to island 1.}
\acrodef{im2}{mirrors the naturally inspired quality that although two populations have the same physical separation, it may be easier to migrate in one direction than the other, i.e. fish migration is easier downstream than upstream.}
\acrodef{digEco}{with the agents, the populations, the agent migration for \acl{DEC}, and the environmental selection pressures provided by the user base, then the union of the habitats creates the Digital Ecosystem}
\acrodef{archComTop}{many strongly connected clusters (communities), called {sub-networks} (quasi-complete graphs), with a few connections between these clusters (communities). Graphs with this topology have a very high clustering coefficient and small characteristic path lengths}
\acrodef{similarCap}{requests are evaluated on separate {islands} (populations), and so adaptation is accelerated by the sharing of solutions between evolving populations (islands), because they are working to solve similar requests (problems).}
\acrodef{picUser}{will formulate queries to the Digital Ecosystem by creating a request as a {semantic description}, like those being used and developed in \acp{SOA}}
\acrodef{picUserReq}{A population is then instantiated in the user's habitat in response to the user's request, seeded from the agents available at their habitat.}
\acrodef{urlCapUnifrom}{The {observed} frequencies of the application (agent aggregation) size mostly matched the {expected} frequencies}
\acrodef{urlCapGaussian}{The {observed} frequencies of the application (agent aggregation) size matched the {expected} frequencies with only minor variations}
\acrodef{urlpower}{The {observed} frequencies of the application (agent aggregation) size matched the {expected} frequencies with some variation}
\acrodef{urvunifromCap}{The {observed} frequencies for the number of agent attributes mostly matched the {expected} frequencies}
\acrodef{urvgaussianCap}{The {observed} frequencies for the number of agent attributes again followed the {expected} frequencies, but there was variation}
\acrodef{urvpowerCap}{The {observed} frequencies for the number of agent attributes also followed the {expected} frequencies, but there was variation}
\acrodef{im1}{are different probabilities of going from island 1 to island 2, as there is of going from island 2 to island 1.}
\acrodef{im2}{mirrors the naturally inspired quality that although two populations have the same physical separation, it may be easier to migrate in one direction than the other, i.e. fish migration is easier downstream than upstream.}
\acrodef{digEco}{with the agents, the populations, the agent migration for \acl{DEC}, and the environmental selection pressures provided by the user base, then the union of the habitats creates the Digital Ecosystem}
\acrodef{archComTop}{many strongly connected clusters (communities), called {sub-networks} (quasi-complete graphs), with a few connections between these clusters (communities). Graphs with this topology have a very high clustering coefficient and small characteristic path lengths}
\acrodef{similarCap}{requests are evaluated on separate {islands} (populations), and so adaptation is accelerated by the sharing of solutions between evolving populations (islands), because they are working to solve similar requests (problems).}
\acrodef{picUser}{will formulate queries to the Digital Ecosystem by creating a request as a {semantic description}, like those being used and developed in \acp{SOA}}
\acrodef{picUserReq}{A population is then instantiated in the user's habitat in response to the user's request, seeded from the agents available at their habitat.}
\acrodef{urlCapUnifrom}{The {observed} frequencies of the application (agent aggregation) size mostly matched the {expected} frequencies}
\acrodef{urlCapGaussian}{The {observed} frequencies of the application (agent aggregation) size matched the {expected} frequencies with only minor variations}
\acrodef{urlpower}{The {observed} frequencies of the application (agent aggregation) size matched the {expected} frequencies with some variation}
\acrodef{urvunifromCap}{The {observed} frequencies for the number of agent attributes mostly matched the {expected} frequencies}
\acrodef{urvgaussianCap}{The {observed} frequencies for the number of agent attributes again followed the {expected} frequencies, but there was variation}
\acrodef{urvpowerCap}{The {observed} frequencies for the number of agent attributes also followed the {expected} frequencies, but there was variation}
\acrodef{im1}{are different probabilities of going from island 1 to island 2, as there is of going from island 2 to island 1.}
\acrodef{im2}{mirrors the naturally inspired quality that although two populations have the same physical separation, it may be easier to migrate in one direction than the other, i.e. fish migration is easier downstream than upstream.}
\acrodef{digEco}{with the agents, the populations, the agent migration for \acl{DEC}, and the environmental selection pressures provided by the user base, then the union of the habitats creates the Digital Ecosystem}
\acrodef{archComTop}{many strongly connected clusters (communities), called {sub-networks} (quasi-complete graphs), with a few connections between these clusters (communities). Graphs with this topology have a very high clustering coefficient and small characteristic path lengths}
\acrodef{similarCap}{requests are evaluated on separate {islands} (populations), and so adaptation is accelerated by the sharing of solutions between evolving populations (islands), because they are working to solve similar requests (problems).}
\acrodef{picUser}{will formulate queries to the Digital Ecosystem by creating a request as a {semantic description}, like those being used and developed in \acp{SOA}}
\acrodef{picUserReq}{A population is then instantiated in the user's habitat in response to the user's request, seeded from the agents available at their habitat.}
\acrodef{urlCapUnifrom}{The {observed} frequencies of the application (agent aggregation) size mostly matched the {expected} frequencies}
\acrodef{urlCapGaussian}{The {observed} frequencies of the application (agent aggregation) size matched the {expected} frequencies with only minor variations}
\acrodef{urlpower}{The {observed} frequencies of the application (agent aggregation) size matched the {expected} frequencies with some variation}
\acrodef{urvunifromCap}{The {observed} frequencies for the number of agent attributes mostly matched the {expected} frequencies}
\acrodef{urvgaussianCap}{The {observed} frequencies for the number of agent attributes again followed the {expected} frequencies, but there was variation}
\acrodef{urvpowerCap}{The {observed} frequencies for the number of agent attributes also followed the {expected} frequencies, but there was variation}

\begin{document}

\title{Computing of Applied Digital Ecosystems}

\numberofauthors{2} 
\author{
\alignauthor
Gerard Briscoe\\
       \affaddr{Digital Ecosystems Lab}\\
       \affaddr{Department of Media and Communications}\\
       \affaddr{London School of Economics}\\
       \affaddr{United Kingdom}\\
       \email{g.briscoe@lse.ac.uk}
\alignauthor
Philippe De Wilde\\
       \affaddr{Intelligent Systems Lab}\\
       \affaddr{Department of Computer Science}\\
       \affaddr{Heriot Watt University}\\
       \affaddr{United Kingdom}\\
       \email{pdw@hw.ac.uk}
}

\maketitle
\begin{abstract}
A primary motivation for our research in \emph{digital ecosystems} is the desire to exploit the self-organising properties of biological ecosystems. Ecosystems are thought to be robust, scalable architectures that can automatically solve complex, dynamic problems. However, the computing technologies that contribute to these properties have not been made explicit in \emph{digital ecosystems} research. Here, we discuss how different computing technologies can contribute to providing the necessary self-organising features, including \acp{MAS}, \acp{SOA}, and \ac{DEC}. The potential for exploiting these properties in \emph{digital ecosystems} is considered, suggesting how several key features of biological ecosystems can be exploited in Digital Ecosystems, and discussing how mimicking these features may assist in developing robust, scalable self-organising architectures. An example architecture, the Digital Ecosystem, is considered in detail. The Digital Ecosystem is then measured experimentally through simulations, considering the self-organised diversity of its evolving agent populations relative to the user request behaviour.

\end{abstract}



\section{Introduction}

Digital Ecosystems are distributed adaptive open socio-technical systems, with properties of self-organisation, scalability and sustainability, inspired by natural ecosystems \cite{thesis}, and are emerging as a novel approach to the catalysis of sustainable regional development driven by \acp{SME}. Digital Ecosystems aim to help local economic actors become active players in globalisation, valorising their local culture and vocations, and enabling them to interact and create value networks at the global level \cite{dini2008bid}.

We have previously considered the biological inspiration for the technical component of Digital Ecosystems \cite{de07oz, dbebkpub}, and we will now consider the relevant computing technologies. Based on our understanding of biological ecosystems in the context of Digital Ecosystems \cite{bionetics}, we will now introduce fields from the domain of computer science relevant to the creation of Digital Ecosystems. As we are interested in the digital counterparts for the behaviour and constructs of biological ecosystems, instead of simulating or emulating such behaviour or constructs, we will consider what parallels can be drawn. We will start by considering \acp{MAS} to explore the references to \emph{agents} and \emph{migration} \cite{de07oz, dbebkpub}; followed by evolutionary computing and \acp{SOA} for the references to \emph{evolution} and \emph{self-organisation} \cite{de07oz, dbebkpub}.

The value of creating parallels between biological and computer systems varies substantially depending on the behaviours or constructs being compared, and sometimes cannot be done so convincingly. For example, both have mechanisms to ensure data integrity. In computer systems, that integrity is absolute, data replication which introduces even the most minor change is considered to have failed, and is supported by mechanisms such as the \acl{MD5} \cite{rivest1992rmm}. While in biological systems, the genetic code is transcribed with a remarkable degree of fidelity; there is, approximately, only one unforced error per one hundred bases copied \cite{Kunkel2004}. There are also elaborate proof-reading and correction systems, which in evolutionary terms are highly conserved \cite{Kunkel2004}. In this example establishing a parallel is infeasible, despite the relative similarity in function, because the operational control mechanisms in biological and computing systems are radically different, as are the aims and purposes. This is a reminder that considerable finesse is required when determining parallels, or when using existing ones.

\section{Multi-Agent Systems}

A \emph{software agent} is a piece of software that acts, for a user in a relationship of \emph{agency}, autonomously in an environment to meet its designed objectives. A \ac{MAS} is a system composed of several \emph{software agents}, collectively capable of reaching goals that are difficult to achieve by an individual agent or monolithic system. Conceptually, there is a strong parallel between the software agents of a \ac{MAS} and the agent-based models of a biological ecosystem \cite{Greenetal2006}, despite the lack of evolution and migration in a \ac{MAS}. There is an even stronger parallel to a variant of \acp{MAS}, called \emph{mobile agent systems}, in which the mobility also mirrors the migration in biological ecosystems \cite{moaspaper}.

\tfigure{width=3.25in}{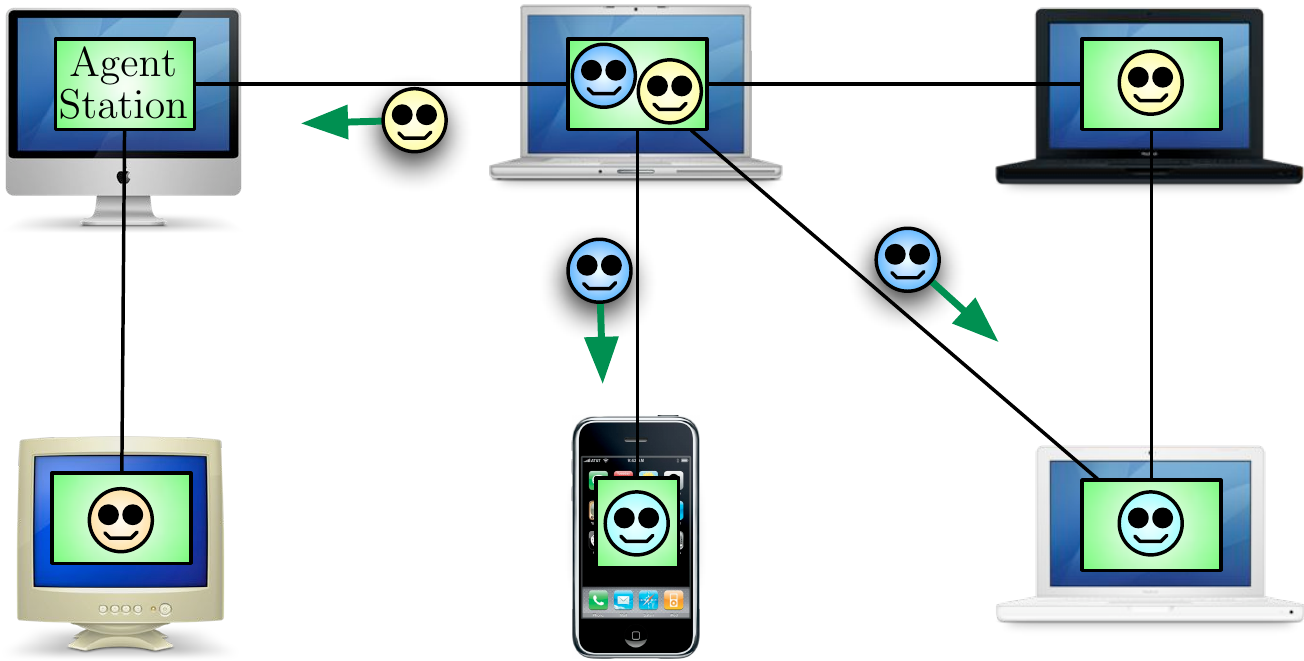}{graffle}{Mobile Agent System}{Visualisation that shows mobile agents as programmes that can migrate from one host to another in a network of heterogeneous computer systems and perform a task specified by its owner \cite{agentStation}.}{-4mm}{}{}{}

The term \emph{mobile agent} contains two separate and distinct concepts: mobility and agency \cite{rothermel1998ma}. Hence, mobile agents are software agents capable of movement within a network \cite{moaspaper}. The mobile agent paradigm proposes to treat a network as multiple agent-\emph{friendly} environments and the agents as programmatic entities that move from location to location, performing tasks for users. So, on each host they visit mobile agents need software which is responsible for their execution, providing a safe execution environment \cite{moaspaper}.

Generally, there are three types of design for mobile agent systems \cite{moaspaper}: (1) using a specialised operating system, (2) as operating system services or extensions, or (3) as application software. The third approach builds mobile agent systems as specialised application software that runs on top of an operating system, to provide for the mobile agent functionality, with such software being called an \emph{agent station} \cite{agentStation}. In this last approach, each agent station hides the vendor-specific aspects of its host platform, and offers standardised services to visiting agents. Services include access to local resources and applications; for example, web servers or \emph{web services}, the local exchange of information between agents via message passing, basic security services, and the creation of new agents \cite{agentStation}. Also, the third approach is the most platform-agnostic, and is visualised in Figure \ref{mobileAgents}.

\section{Service-Oriented Architectures}

To evolve high-level software components in Digital Ecosystems, we propose taking advantage of the \emph{native} method of software advancement, human developers, and the use of \emph{evolutionary computing} for \emph{combinatorial optimisation} of the available software services. This involves treating developer-produced software services as the functional building blocks, as the base unit in a genetic-algorithms-based process. Our approach to evolving high-level software applications requires a modular reusable paradigm to software development, such as \acp{SOA}. \acp{SOA} are the current state-of-the-art approach, being the current iteration of interface/component-based design from the 1990s, which was itself an iteration of event-oriented design from the 1980s, and before then modular programming from the 1970s \cite{krafzig2004ess}. Service-oriented computing promotes assembling application components into a loosely coupled network of services, to create flexible, dynamic business processes and agile applications that span organisations and computing platforms \cite{papazoglou2003soc}. This is achieved through a \ac{SOA}, an architectural style that guides all aspects of creating and using business processes throughout their life-cycle, packaged as services. This includes defining and provisioning infrastructure that allows different applications to exchange data and participate in business processes, loosely coupled from the operating systems and programming languages underlying the applications. Hence, a \ac{SOA} represents a model in which functionality is decomposed into distinct units (services), which can be distributed over a network, and can be combined and reused to create business applications \cite{papazoglou2003soc}.

A \acs{SOA} depends upon service-orientation as its fundamental design principle. In a \acs{SOA} environment, independent services can be accessed without knowledge of their underlying platform implementation. Services reflect a \emph{service-oriented} approach to programming that is based on composing applications by discovering and invoking network-available services to accomplish some task. This approach is independent of specific programming languages or operating systems, because the services communicate with each other by passing data from one service to another, or by co-ordinating an activity between two or more services \cite{papazoglou2003soc}. So, the concepts of \acsp{SOA} are often seen as built upon, and the development of, the concepts of modular programming and distributed computing \cite{krafzig2004ess}.

\acsp{SOA} allow for an information system architecture that enables the creation of applications that are built by combining loosely coupled and interoperable services. They typically implement functionality most people would recognise as a service, such as filling out an online application for an account, or viewing an online bank statement \cite{krafzig2004ess}. Services are intrinsically unassociated units of functionality, without calls to each other embedded in them. Instead of services embedding calls to each other in their source code, protocols are defined which describe how services can talk to each other, in a process known as orchestration, to meet new or existing business system requirements. This is allowing an increasing number of third-party software companies to offer software services, such that \acs{SOA} systems will come to consist of such third-party services combined with others created in-house, which has the potential to spread costs over many users and uses, and promote standardisation both in and across industries. For example, the travel industry now has a well-defined, and documented, set of both services and data, sufficient to allow any competent software engineer to create travel agency software using entirely off-the-shelf software services \cite{cardoso2005isw}. 

The vision of \acsp{SOA} assembling application components from a loosely coupled network of services, that can create dynamic business processes and agile applications that span organisations and computing platforms. It will be made possible by creating compound solutions that use internal organisational software assets, including enterprise information and legacy systems, and combining these solutions with external components residing in remote networks. The great promise of \acsp{SOA} is that the \emph{marginal cost} of creating the n-th application is virtually zero, as all the software required already exists to satisfy the requirements of other applications. Only their \emph{combination} and \emph{orchestration} are required to produce a new application \cite{modi2008}. The \emph{key} is that the interactions between the \emph{chunks} are not specified within the \emph{chunks} themselves. Instead, the interaction of services (all of whom are hosted by unassociated peers) is specified by users in an ad-hoc way, with the intent driven by newly emergent business requirements \cite{leymann2002wsa}.

The pinnacle of \acs{SOA} interoperability, is the exposing of services on the internet as \emph{web services}. A web service is a specific type of service that is identified by a \ac{URI}, whose service description and transport utilise open Internet standards. Interactions between web services typically occur as \ac{SOAP} calls carrying \ac{XML} data content. The interface descriptions of web services are expressed using the \ac{WSDL} \cite{SOApaper2}. The \ac{UDDI} standard defines a protocol for directory services that contain web service descriptions. \ac{UDDI} enables web service clients to locate candidate services and discover their details. Service clients and service providers utilise these standards to perform the basic operations of \acsp{SOA} \cite{SOApaper2}. Service aggregators can then use the \ac{BPEL} to create new web services by defining corresponding compositions of the interfaces and internal processes of existing services \cite{SOApaper2}.

\acs{SOA} services inter-operate based on a formal definition (or contract, e.g. \ac{WSDL}) that is independent of the underlying platform and programming language. Service descriptions are used to advertise the service capabilities, interface, behaviour, and quality \cite{SOApaper2}. The publication of such information about available services provides the necessary means for discovery, selection, binding, and composition of services \cite{SOApaper2}. The (expected) behaviour of a service during its execution is described by its behavioural description (for example, as a workflow process). Also, included is a \ac{QoS} description, which publishes important functional and non-functional service quality attributes, such as service metering and cost, performance metrics (response time, for instance), security attributes, integrity (transactional), reliability, scalability, and availability \cite{SOApaper2}. Service clients (end-user organisations that use some service) and service aggregators (organisations that consolidate multiple services into a new, single service offering) utilise \emph{service descriptions} to achieve their objectives \cite{SOApaper2}. One of the most important and continuing developments in \acsp{SOA} is \acf{SWS}, which make use of \emph{semantic descriptions} for service discovery, so that a client can discover the services semantically \cite{cabral2004asw}. 

There are multiple standards available and still being developed for \acsp{SOA}, most notably of recent being REpresentational State Transfer (REST) \cite{singh2005soc}. The software industry now widely implements a thin SOAP/WSDL/UDDI veneer atop existing applications or components that implement the web services paradigm, but the choice of technologies will change with time. Therefore, the fundamentals of \acsp{SOA} and its services are best defined generically, because \acsp{SOA} are technology agnostic and need not be tied to a specific technology \cite{papazoglou2003soc}. Within the current and future scope of the fundamentals of \acsp{SOA}, there is clearly potential to \emph{evolve} complex high-level software applications from the modular services of \acsp{SOA}, instead of the instruction level evolution currently prevalent in genetic programming \cite{overviewGP}.

\section{Distributed Evolutionary Computing}

Having previously suggested evolutionary computing \cite{de07oz, dbebkpub}, and the possibility of it occurring within a distributed environment, not unlike those found in mobile agent systems, leads us to consider a specialised form known as \ac{DEC}. The fact that evolutionary computing manipulates a population of independent solutions actually makes it well suited for parallel and distributed computation architectures \cite{cantupaz1998spg}. The motivation for using parallel or distributed evolutionary algorithms is twofold: first, improving the speed of evolutionary processes by conducting concurrent evaluations of individuals in a population; second, improving the problem-solving process by overcoming difficulties that face traditional evolutionary algorithms, such as maintaining diversity to avoid premature convergence \cite{stender1993pga}. There are several variants of distributed evolutionary computing, leading some to propose a taxonomy for their classification, with there being two main forms \cite{cantupaz1998spg, stender1993pga}:
\begin{itemize}
\item multiple-population/coarse-grained migration/island 
\item single-population/fine-grained diffusion/neighbourhood 
\end{itemize}

In the coarse-grained \emph{island} models \cite{lin1994cgp, cantupaz1998spg}, evolution occurs in multiple parallel sub-populations (islands), each running a local evolutionary algorithm, evolving independently with occasional \emph{migrations} of highly fit individuals among sub-populations. The core parameters for the evolutionary algorithm of the island-models are as follows \cite{lin1994cgp}:
\begin{itemize}
\item number of the sub-populations: 2, 3, 4, more
\item sub-population homogeneity
\begin{itemize}
\item size, crossover rate, mutation rate
\end{itemize}
\item topology of connectivity: ring, star, fully-connected
\item static or dynamic connectivity
\item migration mechanisms: 
\begin{itemize}
\item isolated/synchronous/asynchronous
\item how often migrations occur 
\item which individuals migrate 
\end{itemize}
\end{itemize}

Fine-grained \emph{diffusion} models \cite{stender1993pga} assign one individual per processor. A local neighbourhood topology is assumed, and individuals are allowed to mate only within their neighbourhood, called a \emph{deme}\arxivfootnote{In biology a deme is a term for a local population of organisms of one species that actively interbreed with one another and share a distinct gene pool \cite{devisser2007ese}.}. The demes overlap by an amount that depends on their shape and size, and in this way create an implicit migration mechanism. Each processor runs an identical evolutionary algorithm which selects parents from the local neighbourhood, produces an offspring, and decides whether to replace the current individual with an offspring. However, even with the advent of multi-processor computers, and more recently multi-core processors, which provide the ability to execute multiple threads simultaneously, this approach would still prove impractical in supporting the number of agents necessary to create a Digital Ecosystem. Therefore, we shall further consider the \emph{island} models.

An example island-model \cite{lin1994cgp, cantupaz1998spg} is visualised in Figure \ref{islandModel}, in which there \setCap{are different probabilities of going from island 1 to island 2, as there is of going from island 2 to island 1.}{im1} This allows maximum flexibility for the migration process, and \setCap{mirrors the naturally inspired quality that although two populations have the same physical separation, it may be easier to migrate in one direction than the other, i.e. fish migration is easier downstream than upstream.}{im2} The migration of the \emph{island} models is like the notion of migration in nature, being similar to the metapopulation models of theoretical ecology \cite{levins1969sda}. This model has also been used successfully in the determination of investment strategies in the commercial sector, in a product known as the Galapagos toolkit \cite{galapagos1}. However, all the \emph{islands} in this approach work on exactly the same problem, which makes it less analogous to biological ecosystems in which different locations can be environmentally different \cite{begon96}. We will take advantage of this property later when defining the \acl{EOA} of Digital Ecosystems.

\tfigure{width=2.5in}{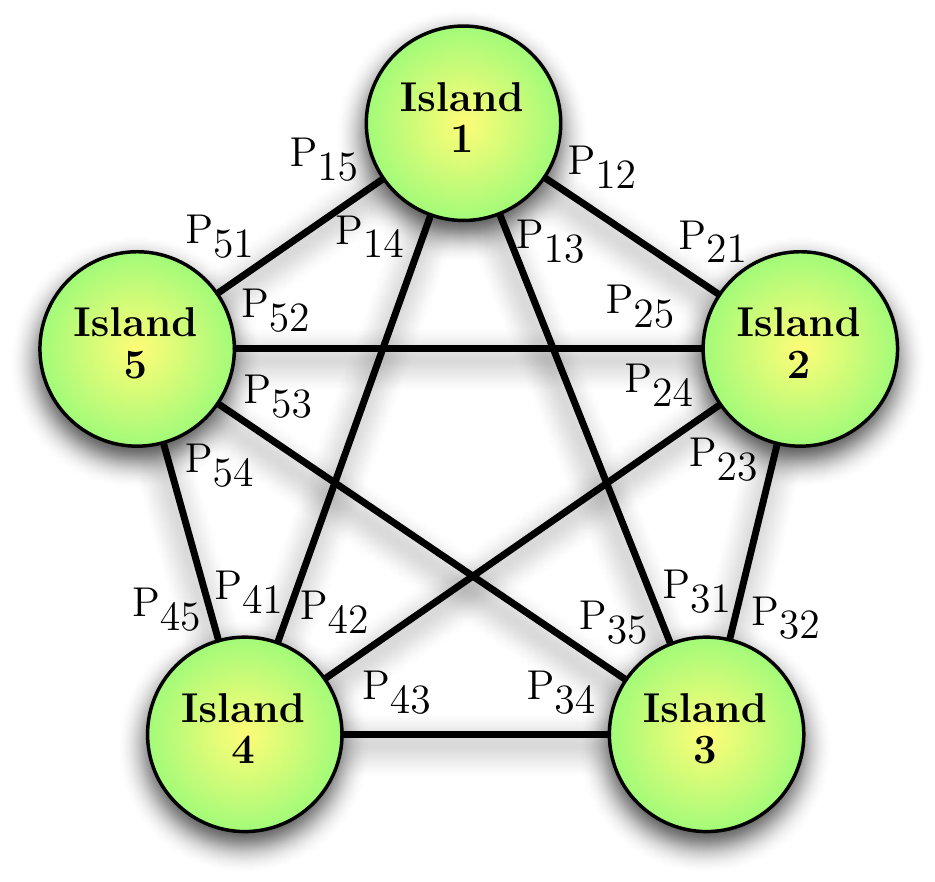}{graffle}{Island-Model of Distributed Evolutionary Computing}{\cite{lin1994cgp, cantupaz1998spg}: There \getCap{im1}}{-6mm}{}{}{-1mm}

\section{The Digital Ecosystem}

Combing these technologies, based on the biological inspiration \cite{bionetics}, the technical component of Digital Ecosystems provide a two-level optimisation scheme inspired by natural ecosystems, in which a decentralised peer-to-peer network forms an underlying tier of distributed agents. These agents then feed a second optimisation level based on an evolutionary algorithm that operates locally on single habitats (peers), aiming to find solutions that satisfy locally relevant constraints. The local search is sped up through this twofold process, providing better local optima as the distributed optimisation provides prior sampling of the search space by making use of computations already performed in other peers with similar constraints \cite{javaOne}. So, the Digital Ecosystem supports the automatic combining of numerous agents (which represent services), by their interaction in evolving populations to meet user requests for applications, in a scalable architecture of distributed interconnected habitats. The sharing of agents between habitats ensures the system is scalable, while maintaining a high evolutionary specialisation for each user. The network of interconnected habitats is equivalent to the \emph{abiotic} environment of biological ecosystems \cite{begon96}; combined \setCap{with the agents, the populations, the agent migration for \acl{DEC}, and the environmental selection pressures provided by the user base, then the union of the habitats creates the Digital Ecosystem}{digEco}, which is summarised in Figure \ref{architecture2}. The continuous and varying user requests for applications provide a dynamic evolutionary pressure on the applications (agent aggregations), which have to evolve to better fulfil those user requests, and without which there would be no driving force to the evolutionary self-organisation of the Digital Ecosystem.

\tfigure{width=3.33in}{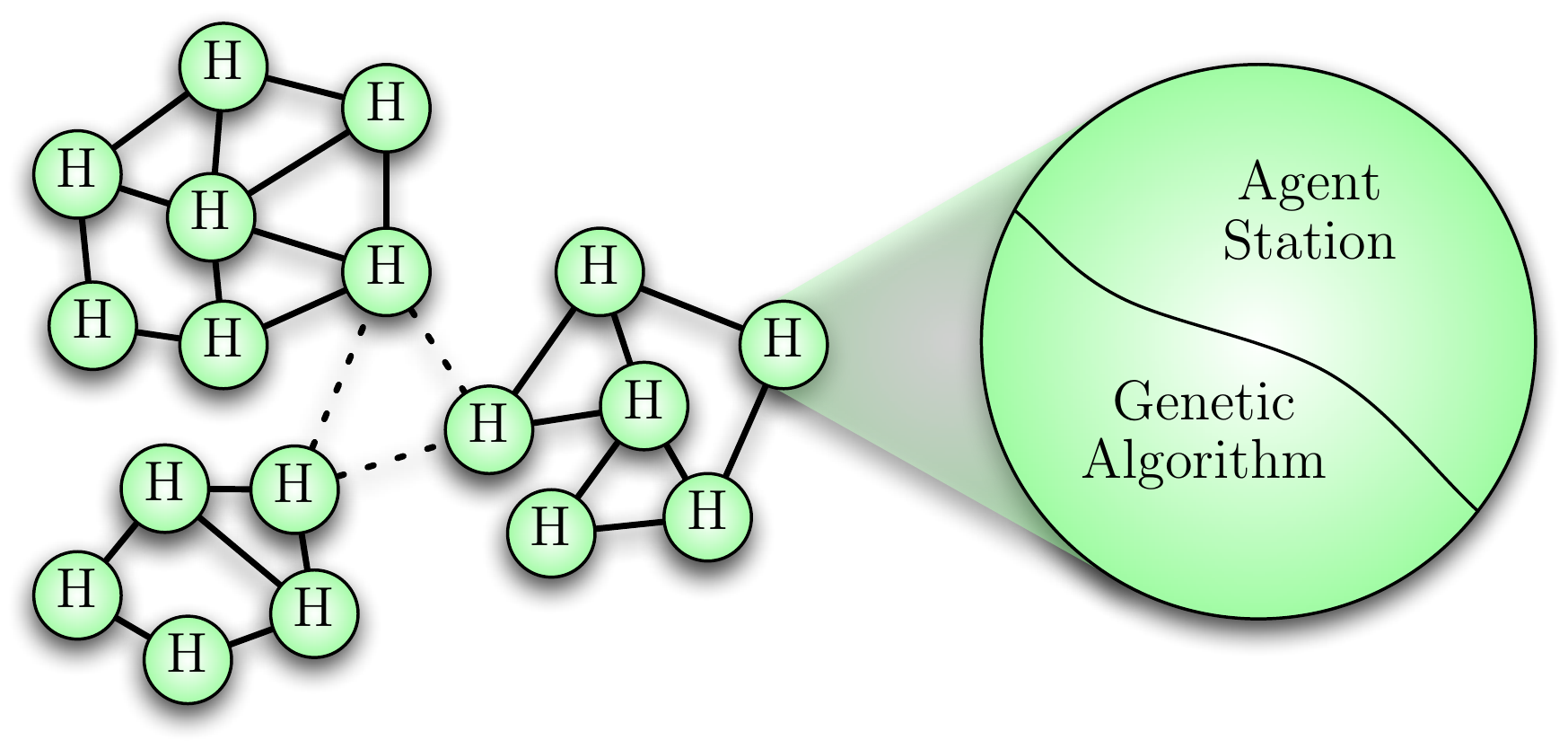}{graffle}{Digital Ecosystem}{Optimisation architecture in which agents (representing services) travel along the P2P connections; in every node (habitat) local optimisation is performed through an evolutionary algorithm, where the search space is determined by the agents present at the node.}{-9mm}{}{}{}

\tfigure{width=3.33in}{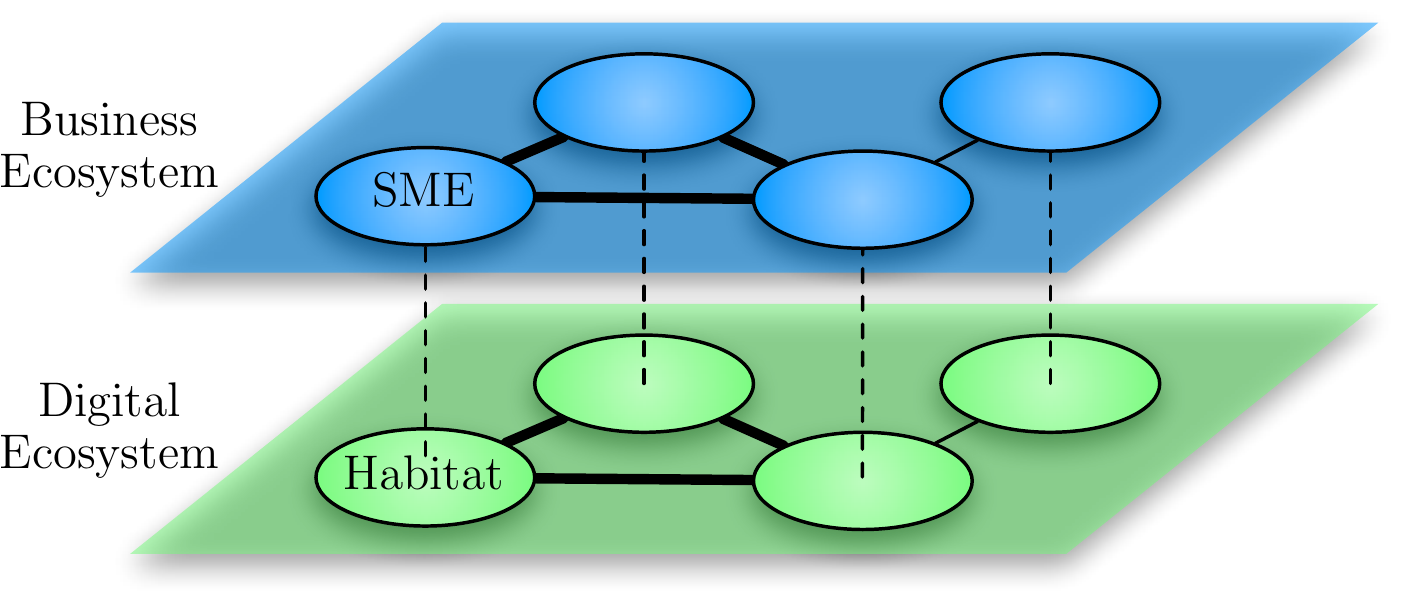}{graffle}{Digital Business Ecosystem}{Business ecosystem, network of \acp{SME} \cite{moore1996}, using the Digital Ecosystem. The habitat clustering will therefore be parallel to the business sector communities.}{-9mm}{!b}{}{}

If we consider an example user base for the Digital Ecosystem, the use of \acp{SOA} in its definition means that \acf{B2B} interaction scenarios \cite{krafzig2004ess} lend themselves to being a potential user base for Digital Ecosystems. So, we can consider a \emph{business ecosystem} of \acf{SME} networks \cite{moore1996}, as a specific class of examples for \ac{B2B} interaction scenarios; and in which the \ac{SME} users are requesting and providing software services, represented as agents in the Digital Ecosystem, to fulfil the needs of their business processes, creating a Digital Business Ecosystem as shown in Figure \ref{DBE}. \acp{SOA} promise to provide potentially huge numbers of services that programmers can combine, via the standardised interfaces, to create increasingly more sophisticated and distributed applications \cite{SOApaper2}. The Digital Ecosystem extends this concept with the automatic combining of available and applicable services, represented by agents, in a scalable architecture, to meet user requests for applications. These agents will recombine and evolve over time, constantly seeking to improve their effectiveness for the user base. From the SME users' point of view the Digital Ecosystem provides a network infrastructure where connected enterprises can advertise and search for services (real-world or software only), putting a particular emphasis on the composability of loosely coupled services and their optimisation to local and regional, needs and conditions. To support these SME users the Digital Ecosystem is satisfying the companies' business requirements by finding the most suitable services or combination of services (applications) available in the network. An application (composition of services) is defined be an agent aggregation (collection) in the habitat network that can move from one peer (company) to another, being hosted only in those where it is most useful in satisfying the \ac{SME} users' business needs.

The agents consist of an \emph{executable component} and an \emph{ontological description}. So, the Digital Ecosystem can be considered a \ac{MAS} which uses \emph{distributed evolutionary computing} \cite{cantupaz1998spg, stender1993pga} to combine suitable agents in order to meet user requests for applications.

The landscape, in energy-centric biological ecosystems, defines the connectivity between habitats \cite{begon96}. Connectivity of nodes in the digital world is generally not defined by geography or spatial proximity, but by information or semantic proximity. For example, connectivity in a peer-to-peer network is based primarily on bandwidth and information content, and not geography. The island-models of \acl{DEC} use an information-centric model for the connectivity of nodes (\emph{islands}) \cite{lin1994cgp}. However, because it is generally defined for one-time use (to evolve a solution to one problem and then stop) it usually has a fixed connectivity between the nodes, and therefore a fixed topology \cite{cantupaz1998spg}. So, supporting evolution in the Digital Ecosystem, with a multi-objective \emph{selection pressure} (fitness landscape with many peaks), requires a re-configurable network topology, such that habitat connectivity can be dynamically adapted based on the observed migration paths of the agents between the users within the habitat network. Based on the island-models of \acl{DEC} \cite{lin1994cgp}, each connection between the habitats is bi-directional and there is a probability associated with moving in either direction across the connection, with the connection probabilities affecting the rate of migration of the agents. However, additionally, the connection probabilities will be updated by the success or failure of agent migration using the concept of Hebbian learning: the habitats which do not successfully exchange agents will become less strongly connected, and the habitats which do successfully exchange agents will achieve stronger connections. This leads to a topology that adapts over time, resulting in a network that supports and resembles the connectivity of the user base. If we consider a \emph{business ecosystem}, network of \acp{SME}, as an example user base; such business networks are typically small-world networks \cite{white2002nst}. They have \setCap{many strongly connected clusters (communities), called \emph{sub-networks} (quasi-complete graphs), with a few connections between these clusters (communities). Graphs with this topology have a very high clustering coefficient and small characteristic path lengths}{archComTop} \cite{swn1}. So, the Digital Ecosystem will take on a topology similar to that of the user base, as shown in Figure \ref{DBE}.

The novelty of our approach comes from the evolving populations being created in response to \emph{similar} requests. So whereas in the island-models of \acl{DEC} there are multiple evolving populations in response to one request \cite{lin1994cgp}, here there are multiple evolving populations in response to \emph{similar} requests. In our Digital Ecosystems different \setCap{requests are evaluated on separate \emph{islands} (populations), and so adaptation is accelerated by the sharing of solutions between evolving populations (islands), because they are working to solve similar requests (problems).}{similarCap}

The users \setCap{will formulate queries to the Digital Ecosystem by creating a request as a \emph{semantic description}, like those being used and developed in \acp{SOA}}{picUser}, specifying an application they desire and submitting it to their local peer (habitat). This description defines a metric for evaluating the \emph{fitness} of a composition of agents, as a distance function between the \emph{semantic description} of the request and the agents' \emph{ontological descriptions}. \setCap{A population is then instantiated in the user's habitat in response to the user's request, seeded from the agents available at their habitat.}{picUserReq} This allows the evolutionary optimisation to be accelerated in the following three ways: first, the habitat network provides a subset of the agents available globally, which is localised to the specific user it represents; second, making use of applications (agent aggregations) previously evolved in response to the user's earlier requests; and third, taking advantage of relevant applications evolved elsewhere in response to similar requests by other users. The population then proceeds to evolve the optimal application (agent aggregation) that fulfils the user request, and as the agents are the base unit for evolution, it searches the available agent combination space. For an evolved agent aggregation (application) that is executed by the user, it then migrates to other peers (habitats) becoming hosted where it is useful, to combine with other agents in other populations to assist in responding to other user requests for applications.

\section{Simulation and Results}

While we could measure the self-organised \emph{diversity} of individual evolving agent populations, or even take a random sampling, it will be more informative to consider their \emph{collective} self-organised \emph{diversity}. Also, given that the Digital Ecosystem is required to support a range of user behaviour, we can consider the \emph{collective} self-organised \emph{diversity} of the evolving agent populations relative to the \emph{global} user request behaviour. So, when varying a behavioural property of the user requests according to some distribution, we would expect the corresponding property of the evolving agent populations to follow the same distribution. We are not intending to prescribe the expected user behaviour of the Digital Ecosystem, but investigate whether the Digital Ecosystem can adapt to a range of user behaviour. So, we will consider Uniform, Gaussian (Normal) and Power Law distributions for the parameters of the user request behaviour. The Uniform distribution will provide a control, while the Normal (Gaussian) distribution will provide a reasonable assumption for the behaviour of a large group of users, and the Power Law distribution will provide a relatively extreme variation in user behaviour.

So, we simulated the Digital Ecosystem, varying aspects of the user behaviour according to different distributions, and measuring the related aspects of the evolving agent populations. This consisted of a mechanism to vary the user request properties of \emph{length} and \emph{modularity} (number of attributes per atomic service), according to Uniform, Gaussian (normal) and Power Law distributions, and a mechanism to measure the corresponding application (agent aggregation) properties of \emph{size} and \emph{number of attributes per agent}. For statistical significance each scenario (experiment) will be averaged from ten thousand simulation runs. We expect it will be obvious whether the \emph{observed} behaviour of the Digital Ecosystem matches the \emph{expected} behaviour from the user base. Nevertheless, we will also implement a chi-squared ($\chi^2$) test to determine if the observed behaviour (distribution) of the agent aggregation properties matches the expected behaviour (distribution) from the user request properties.

\subsection{User Request Length}

We started by varying the \emph{user request length} according to the available distributions, expecting the size of the corresponding applications (agent aggregations) to be distributed according to the length of the user requests, i.e. the longer the user request, the larger the agent aggregation needed to fulfil it.

We first applied the Uniform distribution as a control, and graphed the results in Figure \ref{urluniform}. \setCap{The \emph{observed} frequencies of the application (agent aggregation) size mostly matched the \emph{expected} frequencies}{urlCapUnifrom}, which was \emph{confirmed} by a $\chi^2$ test; with a \emph{null hypothesis} of \emph{no significant difference} and \emph{sixteen degrees of freedom}, the $\chi^2$ value was 2.588, below the critical 0.95 $\chi^2$ value of 7.962. 

We then applied the Gaussian distribution as a reasonable assumption for the behaviour of a large group of users, and graphed the results in Figure \ref{urlgaussian}. \setCap{The \emph{observed} frequencies of the application (agent aggregation) size matched the \emph{expected} frequencies with only minor variations}{urlCapGaussian}, which was confirmed by a $\chi^2$ test; with a \emph{null hypothesis} of \emph{no significant difference} and \emph{sixteen degrees of freedom}, the $\chi^2$ value was 2.102, below the critical 0.95 $\chi^2$ value of 7.962.

Finally, we applied the Power Law distribution to represent a relatively extreme variation in user behaviour, and graphed the results in Figure \ref{urlpower}. \setCap{The \emph{observed} frequencies of the application (agent aggregation) size matched the \emph{expected} frequencies with some variation}{urlpower}, which was confirmed by a $\chi^2$ test; with a \emph{null hypothesis} of \emph{no significant difference} and \emph{sixteen degrees of freedom}, the $\chi^2$ value was 5.048, below the critical 0.95 $\chi^2$ value of 7.962.

There were a couple of minor discrepancies, similar to all the experiments. First, there were a small number of \emph{individual} agents at the thousandth time step, caused by the typical user behaviour of continuously creating new services (agents). Second, while the chi-squared tests confirmed that there was no significant difference between the \emph{observed} and \emph{expected} frequencies of the application (agent aggregation) size, there was still a \emph{bias} to larger applications (solutions). Evident visually in the graphs of the experiments, and evident numerically in the chi-squared test of the Power Law distribution experiment as it favoured shorter agent-sequences. The cause of this \emph{bias} was most likely some aspect of \emph{bloat}\arxivfootnote{When variable length representations of solutions are used, a well-known phenomenon arises, called \emph{bloat}, in which the individuals of an evolving population tend to grow in size without gaining any additional advantage \cite{langdon1997fcb}.} not fully controlled.

\tfigure{width=3.33in}{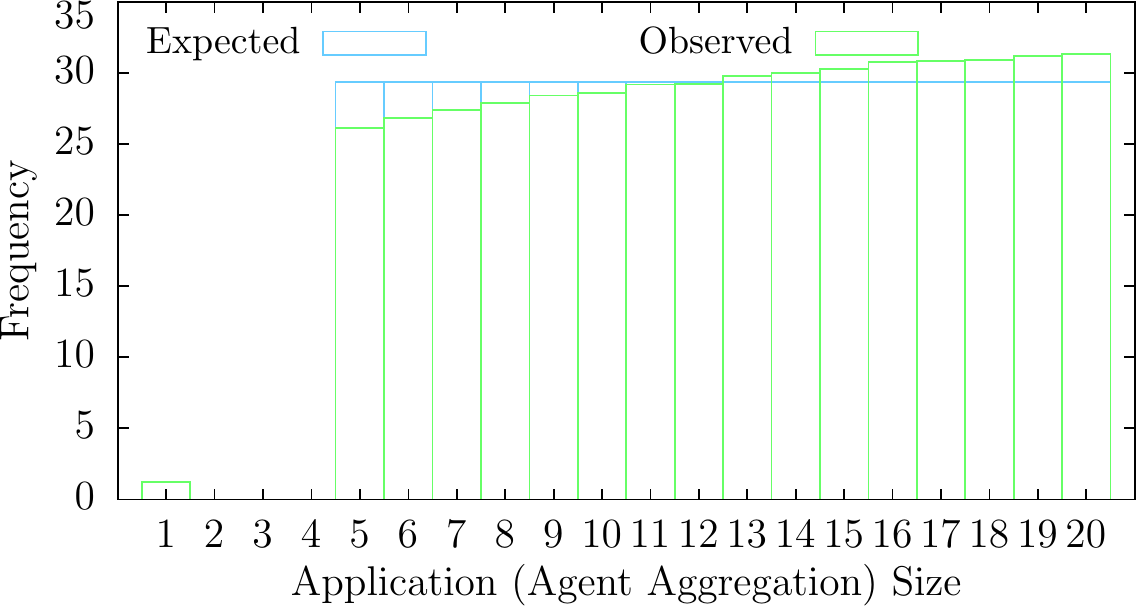}{graph}{Graph of Uniformly Distributed Agent-Sequence Length Frequencies}{\getCap{urlCapUnifrom}.}{-7mm}{}{}{}

\tfigure{width=3.33in}{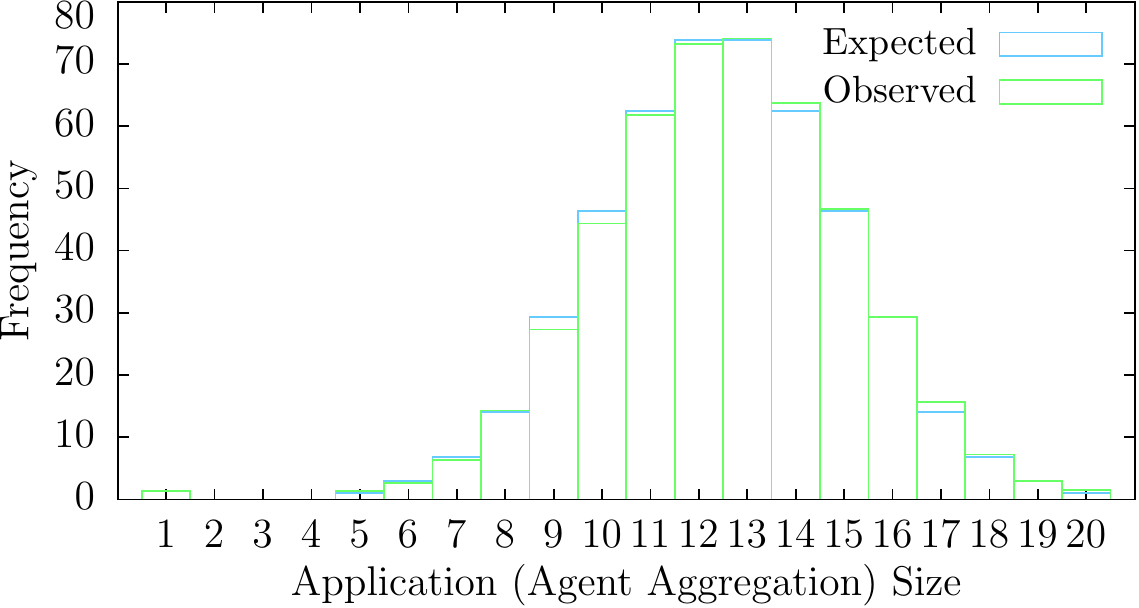}{graph}{Graph of Gaussian Distributed Agent-Sequence Length Frequencies}{\getCap{urlCapGaussian}.}{-7mm}{}{}{}

\tfigure{width=3.33in}{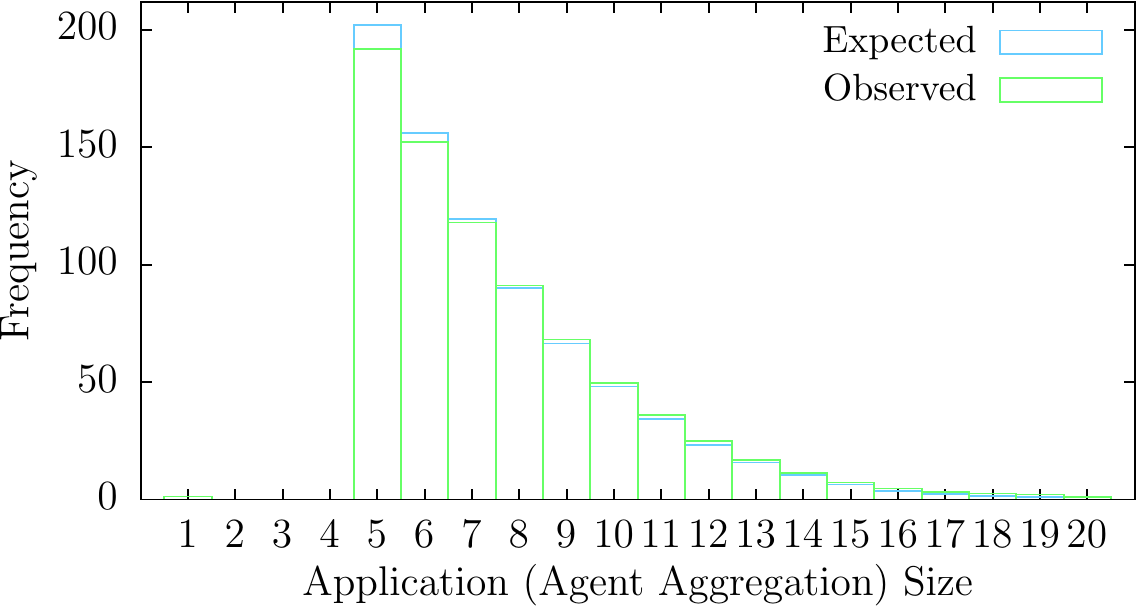}{graph}{Graph of Power Law Distributed Agent-Sequence Length Frequencies}{\getCap{urlpower}.}{-7mm}{}{}{}

\pagebreak\subsection{User Request Modularity}

Next, we varied the \emph{user request modularity} (number of attributes per atomic service) according to the available distributions, expecting the \emph{sophistication} of the agents to be distributed according to the modularity of the user requests, i.e. the more complicated (in terms of modular non-reducible tasks) the user request, the more sophisticated (in terms of the number of attributes) the agents needed to fulfil it.

We first applied the Uniform distribution as a control, and graphed the results in Figure \ref{urvuniform}. \setCap{The \emph{observed} frequencies for the number of agent attributes mostly matched the \emph{expected} frequencies}{urvunifromCap}, which was confirmed by a $\chi^2$ test; with a \emph{null hypothesis} of \emph{no significant difference} and \emph{ten degrees of freedom}, the $\chi^2$ value was 1.049, below the critical 0.95 $\chi^2$ value of 3.940.

\tfigure{width=3.33in}{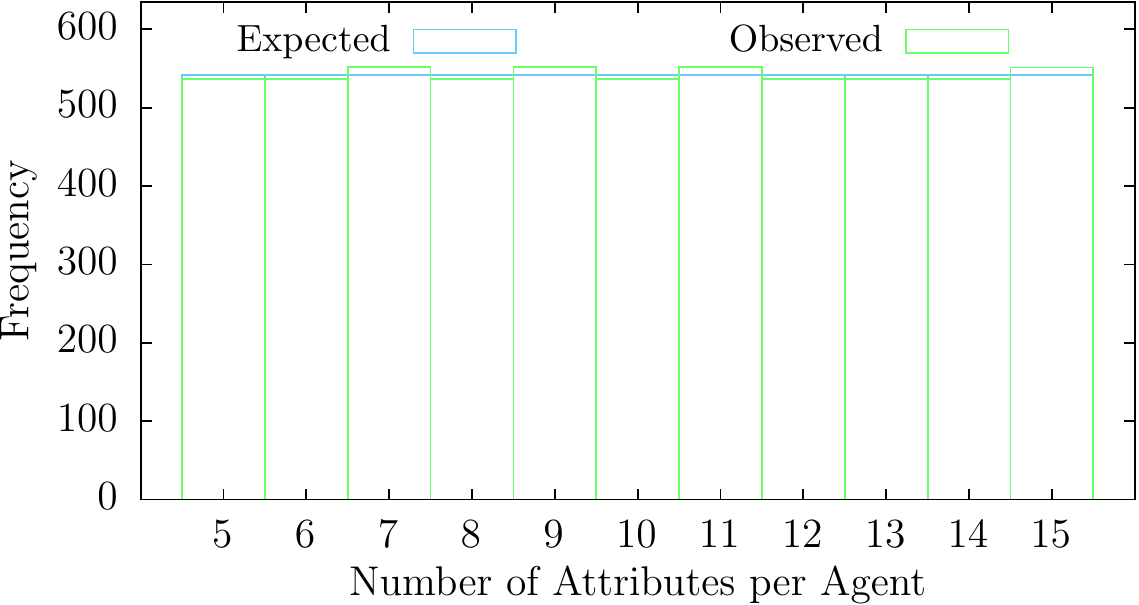}{graph}{Graph of Uniformly Distributed Agent Attribute Frequencies}{\getCap{urvunifromCap}}{-8mm}{}{}{}

\tfigure{width=3.33in}{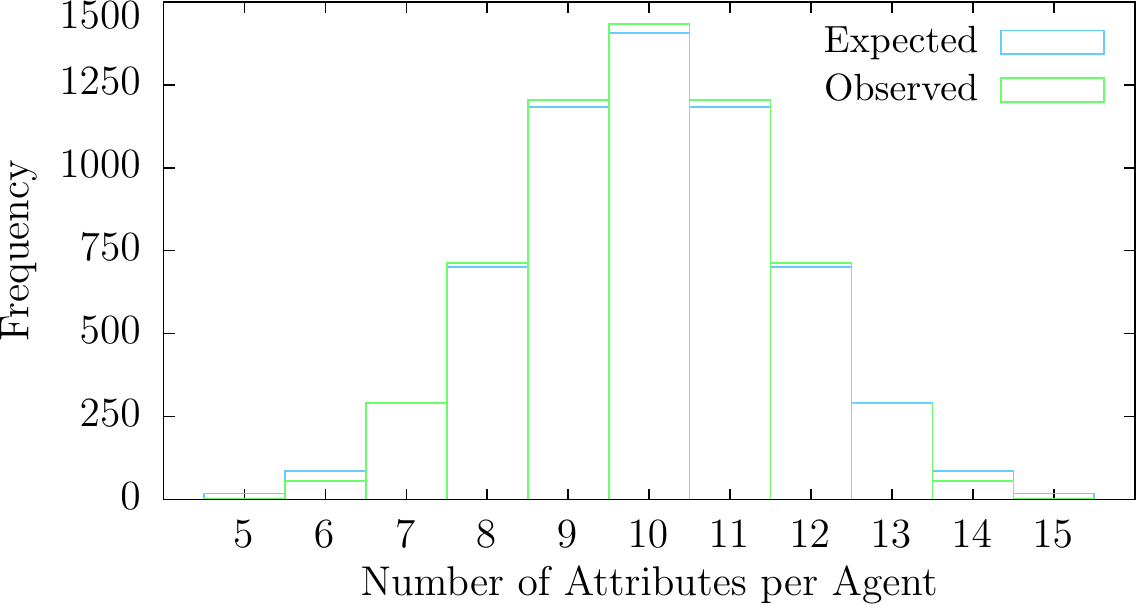}{graph}{Graph of Gaussian Distributed Agent Attribute Frequencies}{\getCap{urvgaussianCap}.}{-7mm}{}{}{}

\label{divmodexp}

\tfigure{width=3.33in}{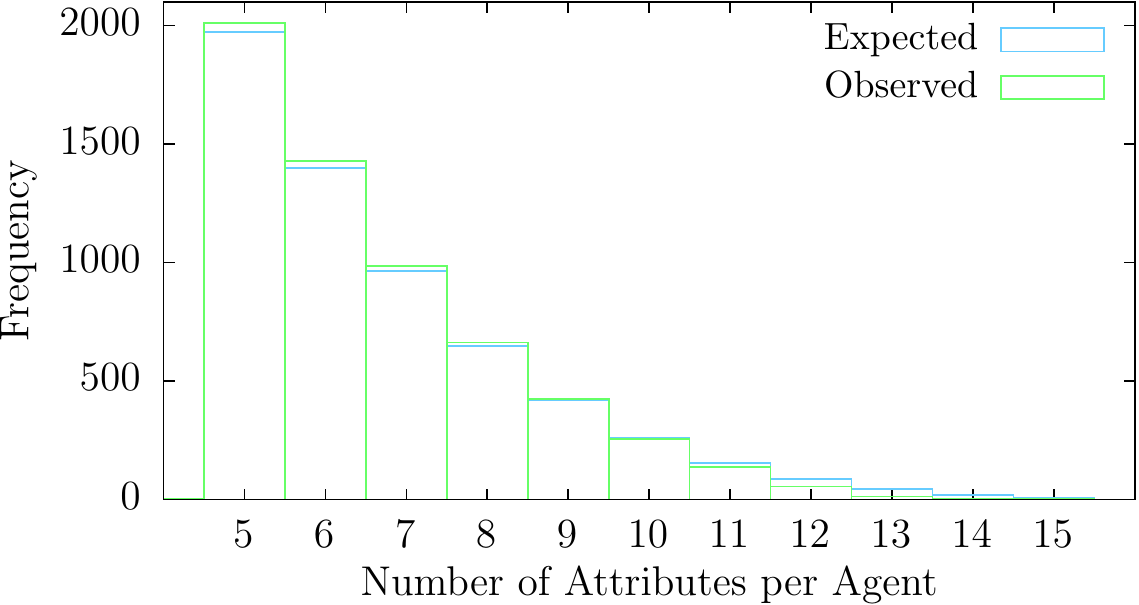}{graph}{Graph of Power Law Distributed Agent Attribute Frequencies}{\getCap{urvpowerCap}.}{-7mm}{}{}{}

We then applied the Gaussian distribution as a reasonable assumption for the behaviour of a large group of users, and graphed the results in Figure \ref{urvgaussian}. \setCap{The \emph{observed} frequencies for the number of agent attributes again followed the \emph{expected} frequencies, but there was variation}{urvgaussianCap} which led to a failed $\chi^2$ test; with a \emph{null hypothesis} of \emph{no significant difference} and \emph{ten degrees of freedom}, the $\chi^2$ value was 50.623, not below the critical 0.95 $\chi^2$ value of 3.940.

Finally, we applied the Power Law distribution to represent a relatively extreme variation in user behaviour, and graphed the results in Figure \ref{urvpower}. \setCap{The \emph{observed} frequencies for the number of agent attributes also followed the \emph{expected} frequencies, but there was variation}{urvpowerCap} which led to a failed $\chi^2$ test; with a \emph{null hypothesis} of \emph{no significant difference} and \emph{ten degrees of freedom}, the $\chi^2$ value was 61.876, not below the critical 0.95 $\chi^2$ value of 3.940.

In all the experiments the \emph{observed} frequencies of the number of agent attributes followed the \emph{expected} frequencies, with some variation in two of the experiments. Collectively, the experimental results confirm that the self-organised \emph{diversity} of the evolving agent populations is relative to the \emph{selection pressures} of the user base, which was confirmed statistically for most of the experiments. While the minor experimental failures, in which the Digital Ecosystem responded more slowly than in the other experiments, have shown that there is potential to optimise the Digital Ecosystem, because the evolutionary self-organisation of an ecosystem is a slow process \cite{begon96}, even the accelerated form present in our Digital Ecosystem.

\section{Conclusions}

We have confirmed the fundamentals for a new class of system, Digital Ecosystems, created through combining understanding from theoretical ecology, evolutionary theory, \acp{MAS}, \acl{DEC}, and \acp{SOA}. Digital Ecosystems, where the word \emph{ecosystem} is more than just a metaphor, being the digital counterpart of biological ecosystems, and therefore having their desirable properties, such as scalability and self-organisation. It is a complex system that shows emergent behaviour, being more than the sum of its constituent parts.

The ever-increasing challenge of software complexity in creating progressively more sophisticated and distributed applications, makes the design and maintenance of efficient and flexible systems a growing challenge \cite{newsArticle3}, for which current software development techniques have hit a \emph{complexity wall} \cite{lyytinen2001nwn}. In response we have created Digital Ecosystems, the digital counterparts of biological ecosystems, possessing their properties of self-organisation, scalability and sustainability \cite{Levin}; \aclp{EOA} that overcome the challenge by automating the search for new algorithms in a scalable architecture, through the evolution of software services in a distributed network.

\section{Acknowledgments}

We would like to thank for helpful discussions Paolo Dini of the London School of Economics and Political Science. This work was supported by the EU-funded \ac{OPAALS} Network of Excellence (NoE), Contract No. FP6/IST-034824.

\pagebreak
\bibliographystyle{abbrv}
\bibliography{references} 
\end{document}